\documentclass[11pt]{article}

\usepackage[preprint]{acl}

\usepackage{times}
\usepackage{latexsym}

\usepackage[T1]{fontenc}

\usepackage[utf8]{inputenc}

\usepackage{microtype}

\usepackage{inconsolata}

\usepackage{graphicx}
\usepackage{booktabs}
\usepackage{xspace}
\usepackage[table]{xcolor}
\usepackage{array}

\definecolor{CustomGreen}{HTML}{1DA35F}
\newcommand{\second}[1]{\underline{#1}}

\newcommand{\ours}{{\textsc{GRAFT}}\xspace}
\newcommand{\nbench}{six\xspace}
\title{GRAFT: Execution-Guided Genetic Recombination for Agent Design}

\author{
  \textbf{Sevda Ebrahimi\textsuperscript{1,2}},
  \textbf{Mohammed Saidul Islam\textsuperscript{2}},
  \textbf{Xiaodan Zhu\textsuperscript{1,3}},
  \textbf{Elham Dolatabadi\textsuperscript{1,2}}
  \\
  \\
  \textsuperscript{1}York University,
  \textsuperscript{2}Vector Institute, 
  \textsuperscript{3}Queen's University,
  \\
  \small{
    \textbf{Correspondence:}
    \href{mailto:email@domain}{sevdaebr@yorku.ca}
  }
}

\begin{document}
\maketitle

\begin{figure*}[t]
  \centering
  \includegraphics[width=0.75\textwidth]
    {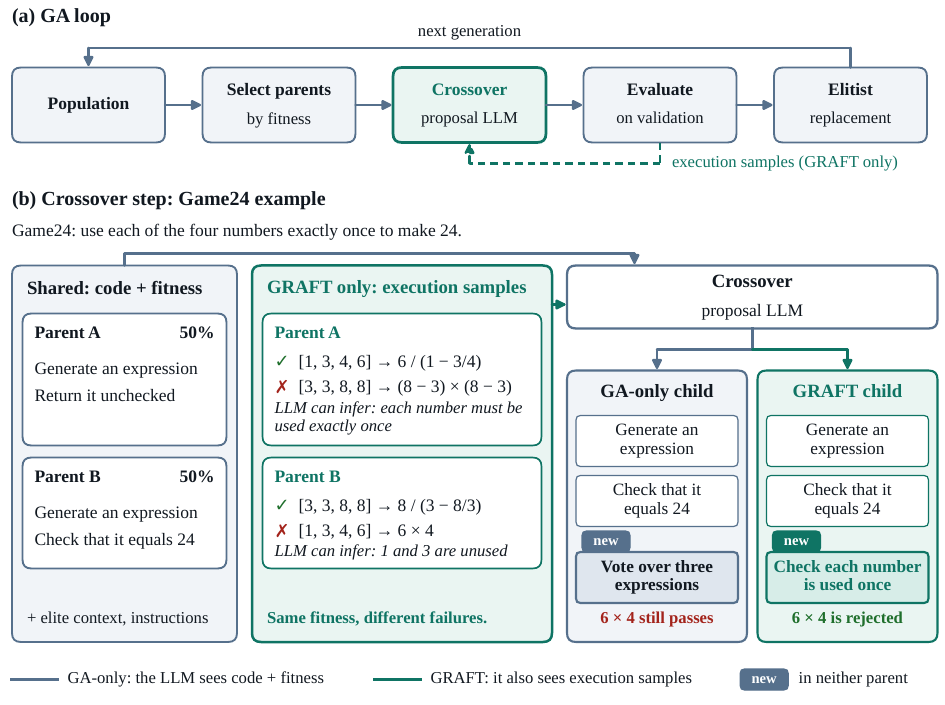}
  \caption{(a)~The GA loop shared by GA-only search and GRAFT. GRAFT also
    passes execution samples from the evaluation step to the crossover step.
    (b)~An illustrative crossover on Game24. Both parents score 50\%, but their execution samples show
    different failures. Both children inherit Parent~B's check that the value
    equals 24, and each adds one step that neither parent has (\emph{new};
    integrated mutation). Without samples, the new step and combination are untargeted, so the
    answer $6\times4$ for [1,\,3,\,4,\,6] still passes. With samples, the child
    adds a check that each number is used once, which rejects it.}
  \label{fig:gradient-guided-crossover}
\end{figure*}

\begin{abstract}

Automated agent design searches for executable architectures that improve LLM-based agent performance, with evolutionary methods using language-model crossover to recombine promising mechanisms for task decomposition, model coordination, and verification. This search, however, is typically guided by aggregate fitness alone, which hides the instance-level successes and failures behind each score. We introduce Execution-Guided Genetic Recombination for Agent Design (\ours), which uses aggregate fitness for selection and survival and parent-specific execution evidence to guide semantic crossover with integrated mutation. Its feedback harness reuses bounded samples of successful and failed runs from fitness evaluation, withholding reference answers and requiring no additional task executions. We evaluate \ours{} across \nbench reasoning and programming settings using open-weight optimizer and executor models. It achieves a mean test accuracy of 82.0\%,
exceeding the strongest baseline by 9.8\% points. Controlled ablations show that adding execution evidence improves test accuracy with initialization, search rules, repair policies, and evaluation budgets held fixed. Our results support the value of parent-specific execution evidence for evolutionary agent design. All code is open-sourced at \url{https://anonymous.4open.science/r/GRAFT-B1EB/}.

\end{abstract}

\section{Introduction}
Architectural decisions, including task decomposition, the orchestration of model calls, and the validation of intermediate outputs, shape the performance of LLM-based agents \citep{bajoria2026language}. Automated agent design therefore searches over executable programs to find architectures that, with a fixed execution model, perform well on unseen instances of a target task \citep{gulli2025agentic}.

Automated search methods, including ADAS \citep{hu2025automated}, AFlow \citep{zhang2025aflow}, and SwarmAgentic \citep{zhang2025swarmagentic}, discover agent architectures through iterative generation and evaluation. Evolutionary approaches can reuse promising mechanisms through language-model crossover, which generates offspring conditioned on multiple parent programs, allowing an LLM to combine and adapt mechanisms expressed in their code \citep{meyerson2024language}. AgentBreeder applies evolutionary variation to agent scaffolds, providing a direct precedent for evolutionary search over agent architectures \citep{rosser2026agentbreeder}.


Evolutionary search is typically guided by aggregate fitness scores, which summarize overall performance without revealing the instance-level successes, failures, and behavioral patterns underlying those scores \citep{guo2023evoprompt, ye2024reevo,meyerson2024language}. Prior work has shown that richer evaluation feedback can guide prompt optimization, workflow refinement, and evolutionary search by providing qualitative, gradient-like signals for improvement \citep{agrawal2026gepa,lee2026meta,lee2025evolving, cheng2024trace,yuksekgonul2025optimizing,pryzant2023automatic,yang2021gradient,liu2026evox}. Yet, its role in guiding semantic crossover and mutation in agent architecture search remains underexplored. 

We investigate this gap by varying the information supplied to LLM-based semantic crossover while holding initialization, search rules, repair policies, and the evaluation budget fixed. We introduce \textbf{Execution-Guided Genetic Recombination for Agent Design (\ours)}, a population-based framework for automated agentic system design. In \ours, aggregate fitness governs parent selection and survival, while an LLM-based variation operator generates offspring through semantic crossover with integrated mutation. Building on Language Model Crossover \citep{meyerson2024language}, the crossover operator receives parent-specific execution evidence supplied by a feedback harness, alongside parent code and fitness (Figure~\ref{fig:gradient-guided-crossover}). We evaluate \ours on \nbench benchmarks spanning abstract reasoning, multilingual mathematics, program synthesis, and arithmetic puzzles, using open-weight optimizer and executor models. With \texttt{Qwen3.5-35B-A3B} as both optimizer and executor, \ours achieves the highest average test accuracy (82.0\%), exceeding the strongest baseline by 9.8 percentage points. On ARC, with a fixed optimizer, \ours achieves the highest mean accuracy across three executors (38.1\%), exceeding the strongest baseline by 4.9 percentage points.

Our contributions are as follows:
\begin{enumerate}

\item \textbf{Execution-Guided Genetic Recombination for Agent Design.}
We introduce \ours, a framework that augments LLM-based semantic crossover with parent-specific execution evidence to guide the combination and revision of executable agent programs.

\item \textbf{A controlled evaluation of architecture search and execution feedback.}
We isolate the role of execution evidence by varying the information supplied to crossover while holding initialization, search rules, repair policies, and the evaluation budget fixed. We assess the performance of \ours across four tasks and \nbench benchmarks and analyze its behavior under different executor models, enabling a systematic comparison with existing automated agent-design methodologies.


\end{enumerate}

\section{Related Work}
\label{sec:related}

\paragraph{Automated agent design.}
Automated designers optimize different aspects of agentic systems.
DSPy optimizes prompts and demonstrations \citep{khattab2024dspy},
while GPTSwarm optimizes node prompts and graph connectivity
\citep{zhuge2024language}. Workflow and architecture search methods
include AFlow's Monte Carlo tree search \citep{zhang2025aflow},
MaAS's learned agentic supernet \citep{zhang2025multi}, and W4S's
reinforcement-trained meta-agent \citep{nie2025weak}. ADAS searches
over executable agent programs using an archive of previous designs
\citep{hu2025automated}, while G{\"o}del Agent and DGM support
self-improvement through agent-code modification
\citep{yin2025godel,zhang2026darwin}. SwarmAgentic maintains a population
of systems and applies PSO-inspired, language-driven updates
\citep{zhang2025swarmagentic}.

\paragraph{LLM-driven evolutionary search.}
LLMs can serve as variation operators within evolutionary search.
ELM uses LLM-generated mutations \citep{lehman2023evolution}, while LMX
generates offspring conditioned on multiple parents
\citep{meyerson2024language}. FunSearch, AlphaEvolve, and ShinkaEvolve
apply LLM-driven evolution to program discovery
\citep{romera2024mathematical,novikov2025alphaevolve,lange2026shinkaevolve}.
Closer to our setting, AgentBreeder evolves executable agent
scaffolds through LLM-based crossover and mutation
\citep{rosser2026agentbreeder}. These approaches motivate combining
population-based search with semantic variation over agent programs.

\paragraph{Feedback-guided optimization.}
Behavioral evidence can complement aggregate fitness in guiding
search. Genetic programming provides precedents through case-wise
performance in lexicase selection \citep{helmuth2014solving}
and program-output semantics in geometric semantic variation
\citep{moraglio2012geometric}. ProTeGi and TextGrad use natural-language
feedback as textual gradients
\citep{pryzant2023automatic,yuksekgonul2025optimizing}, while Trace
combines execution traces with output feedback
\citep{cheng2024trace}. ReEvo supplies reflections on parental
performance to heuristic crossover \citep{ye2024reevo}.
GEPA combines trajectory-based reflection with instance-wise
Pareto selection \citep{agrawal2026gepa}; Mind Evolution uses
evaluator feedback to refine candidate solutions
\citep{lee2025evolving}; and Meta-Harness provides access
to prior programs, scores, and execution traces
\citep{lee2026meta}. Our focus is on using parent-specific execution evidence to guide LMX crossover within a genetic search method for high-performing agent architectures.


\section{Harness-Guided Genetic Agent Design}
\subsection{Problem Formulation}
\label{sec:formulation}

Each candidate agentic system, $S$,  is an executable Python program that takes a task instance as input and returns an answer. The search
space comprises programs specifying the agent's architecture, including
the organization of model calls, intermediate processing, and output
generation. Each candidate is evaluated on a fixed validation set and
assigned a fitness score $J(S)$ reflecting its task performance.

Let $P_t=\{S_i^{(t)}\}_{i=1}^{N}$ denote a population of $N$ candidate agents at iteration $t$. Each candidate is evaluated on a validation set and assigned a fitness score $J(S_i^{(t)})$ reflecting its performance on the target task. The population evolves through iterative selection and variation to discover higher-performing architectures. Search proceeds for at most $n$ iterations, subject to the available design budget and predefined stopping conditions. If search terminates at iteration $T\leq n$, the best-performing agent is selected as:
\begin{equation}
   S_{\mathrm{best}}\in\arg\max_{S\in P_T}J(S).
\label{eq}
\end{equation}

\subsection{Execution-Guided Genetic Recombination for Agent Design}
\label{sec:genetic_search}

Our approach, \ours, combines fitness-based evolutionary selection with
execution-guided semantic crossover within a genetic algorithm. Aggregate fitness determines which
architectures are selected and retained, while parent-specific execution
evidence informs how the LLM constructs offspring. The guidance is
qualitative: the operator uses observed behavior to propose architectural
changes, and fitness is assessed through execution.

\paragraph{Initialization and parent selection.}
The initial population $P_0$ contains $N$ hand-designed agent programs.
Evaluating a program $S$ on the validation set yields its fitness $J(S)$
and per-example execution records $R(S)$. These values are cached together
under the evaluated program's identity. At each iteration $t$, the search
attempts $\lambda$ offspring proposals. Fitness-based tournament selection
chooses two distinct parent programs, $S_a$ and $S_b$, from $P_t$;
each tournament selects a highest-fitness member among $k$ sampled
candidates. Distinct parents may have identical aggregate fitness.

\paragraph{Execution-Feedback Harness.}
For each selected parent, the harness retrieves its cached fitness and execution records to construct a compact representation of its observed performance:
\begin{equation}
    H(S)=\bigl(J(S),E^+(S),E^-(S)\bigr),
    \label{eq:gg-lmx-harness}
\end{equation}
where $E^+(S)$ and $E^-(S)$ denote bounded samples of successful and failed executions, respectively, drawn from $R(S)$. Each record contains the agent's response, correctness outcome, and available evaluator feedback. Successful and failed examples are sampled separately in validation order, subject to the limits specified in the experimental setup. The evidence remains associated with its source parent, allowing the variation operator to distinguish their observed behaviors. Reference answers are withheld from the proposal LLM. Because the harness reuses records collected during fitness evaluation, constructing this feedback requires no additional task executions.

\paragraph{Harness-Guided Semantic Crossover.}

We build on Language Model crossover parent-conditioned generation, in which an LLM produces offspring by recombining information from multiple parent programs. In \ours, the variation model $M_{\mathrm{var}}$ receives both parent programs and their execution-feedback harnesses to generate a complete offspring program.

Let $\mathcal L_t$ denote the elite-population context and $\pi$ the fixed
variation instructions which is specific to each task. Offspring generation is expressed as
\begin{equation}
  \begin{array}{@{}l@{}}
    \widetilde S_c \leftarrow
      \mathrm{GG\mbox{-}LMX}_{M_{\mathrm{var}}}\bigl(\\
      \qquad S_a,S_b,H(S_a),H(S_b),\mathcal L_t;\pi\bigr).
  \end{array}
  \label{eq:gg-lmx-proposal}
\end{equation}
The prompt directs the LLM to preserve or combine useful parental
mechanisms, address observed failure patterns, and introduce architectural
novelty using the elite population as context. Crossover and mutation are
integrated within the same proposal step. Bounded revision passes check
the proposed program's recombination and implementation. Algorithm~\ref{alg:gg-lmx} summarizes the GG-LMX operator.
The complete \ours{} search procedure is presented in Algorithm~\ref{alg:h-gad} in Appendix~D.


\paragraph{Evaluation and Elitist Replacement.}
Each proposal is evaluated using the common validation protocol and the
fixed revision and retry policy. The evaluation routine returns the final
program version $S_c$ together with its fitness and execution records, or
reports an unresolved failure. Fitness and evidence are always associated with the
program version that produced them.

Let $Q_t$ denote the valid, evaluated offspring admitted to replacement.
Because proposal attempts may fail, $|Q_t|\leq\lambda$. The next population
is
\begin{equation}
  P_{t+1}=\mathrm{Top}_N\bigl(P_t\cup Q_t;J\bigr),
  \label{eq:gg-lmx-replacement}
\end{equation}
where $\mathrm{Top}_N$ retains the $N$ highest-fitness candidates,
with ties favoring incumbents. Retained programs keep their cached fitness
and execution evidence. Search ends after at most $n$ iterations, when the
design budget is exhausted, or when the specified stopping condition is
met. It returns a highest-fitness system $S_{\mathrm{best}}$ from the final
population.

\newcounter{gglmxalgorithm}
\newcounter{gglmxline}
\newcommand{\gglmxheading}[1]{%
  \refstepcounter{gglmxalgorithm}%
  \setcounter{gglmxline}{0}%
  \hrule height 0.8pt
  \vspace{2pt}%
  {\raggedright\noindent\textbf{Algorithm \thegglmxalgorithm} #1\par}%
  \vspace{2pt}\hrule height 0.4pt\vspace{3pt}%
}
\newcommand{\gglmxstep}[2]{%
  \stepcounter{gglmxline}%
  {\footnotesize\thegglmxline:} &
  \raggedright\leftskip=#1em\relax #2\tabularnewline
}

\begin{table}[t]
\gglmxheading{\mbox{GG-LMX} Offspring Generation}
\label{alg:gg-lmx}
\noindent\hangindent=1.3em\hangafter=1
\textbf{Require:} Elite context $\mathcal L_t$, proposal LLM
$M_{\mathrm{var}}$, and variation instructions $\pi$.\par
\vspace{2pt}
\noindent\begin{tabular}{@{}r@{\hspace{0.5em}}p{\dimexpr\linewidth-2em\relax}@{}}
\gglmxstep{0}{\textbf{Input:} Parents $S_a,S_b$ and their cached fitness and execution records.}
\gglmxstep{0}{\textbf{Output:} Offspring proposal $\widetilde S_c$.}
\gglmxstep{0}{\textit{// Execution-guided variation}}
\gglmxstep{0}{\textbf{Build harnesses:} Construct $H(S_a)$ and $H(S_b)$ from the cached parent evaluations.}
\gglmxstep{0}{\textbf{Assemble prompt:} Include the parent programs, harness evidence, elite context, and instructions.}
\gglmxstep{0}{\textbf{Crossover and mutation:} Ask $M_{\mathrm{var}}$ to combine parental mechanisms and introduce architectural novelty (integrated mutation).}
\gglmxstep{0}{\textbf{Revise:} Apply bounded revision to check recombination and implementation.}
\gglmxstep{0}{\textbf{Return} $\widetilde S_c$.}
\end{tabular}
\par\vspace{2pt}\hrule height 0.8pt
\end{table}

\section{Experiments}
\label{sec:experiments}

\subsection{Experimental Setup}
\label{sec:experimental_setup}

\paragraph{Tasks.}

We evaluate agentic systems across \nbench benchmarks spanning two broad task categories: reasoning and coding. We use ARC \citep{chollet2019measure} to assess abstract reasoning, MGSM \citep{shi2022language} and Game of 24
\citep{yao2023tree} test multilingual mathematical reasoning and arithmetic
puzzle solving, and SuperGPQA \citep{du2026supergpqa} for domain-specific knowledge and reasoning. We consider two disjoint subsets of SuperGPQA: SuperGPQA-STEM, covering science, engineering, and medicine, and SuperGPQA-Hard-Calc, focusing on challenging calculation-based questions across disciplines. We use MBPP+ \citep{austin2021program,liu2023your} to evaluate code generation with expanded test suites. In addition in the optimizer-variation study, we include CRUXEval-I \citep{gu2024cruxeval} to assess code execution reasoning through input prediction. Further benchmark details are provided in Appendix~\ref{app:benchmarks}.

\begin{table*}[t]
  \centering
  \resizebox{1\textwidth}{!}{%
  \setlength{\tabcolsep}{4pt}
    \begin{tabular}{lc|cc|c|cc|>{\columncolor{blue!8}}c}
    \toprule
    \textbf{Method}
      & \shortstack{\textbf{Abstract}\\\textbf{Reasoning}}
      & \multicolumn{2}{c|}{\textbf{Mathematics}}
      & \shortstack{\textbf{Program}\\\textbf{Synthesis}}
      & \multicolumn{2}{c|}{\textbf{Scientific Reasoning}}
      & \textbf{Overall} \\
    \cmidrule(lr){2-2}
    \cmidrule(lr){3-4}
    \cmidrule(lr){5-5}
    \cmidrule(lr){6-7}
      & \textbf{ARC}
      & \textbf{MGSM}
      & \textbf{Game24}
      & \textbf{MBPP+}
      & \shortstack{\textbf{SuperGPQA}\\\textbf{STEM}}
      & \shortstack{\textbf{SuperGPQA}\\\textbf{Hard-Calc}}
      & \textbf{Average} \\
    \midrule
    \multicolumn{8}{c}{{Hand-designed reference agents}} \\
    \midrule
    CoT \citep{wei2022chain}
      & 19.2 & 12.5 & 88.3 & 75.8 & 43.0 & 32.0 & 45.1 \\
    CoT-SC \citep{wang2022self}
      & 30.8 & 7.8 & 92.2 & 80.5 & 44.5 & 29.7 & 47.6 \\
    Self-Refine \citep{madaan2023self}
      & 32.9 & 11.7 & 91.4 & 78.9 & 60.9 & 57.8 & 55.6 \\
    LLM Debate \citep{du2023improving}
      & 28.7 & 2.6 & \textbf{99.2} & 77.3 & 59.4 & 57.0 & 54.0 \\
    Step-back Abstraction \citep{zheng2024take}
      & 25.2 & 8.6 & \second{96.9} & 66.2 & 58.6 & 46.9 & 50.4 \\
    Quality-Diversity \citep{lu2025intelligent}
      & 29.6 & 3.1 & 91.4 & 76.6 & 59.4 & 55.5 & 52.6 \\
    Role Assignment \citep{xu2023expertprompting}
      & 8.3 & 5.5 & 82.5 & 75.1 & 42.2 & 27.3 & 40.2 \\
    \midrule
    \multicolumn{8}{c}{{Automated agent design}} \\
    \midrule
    ADAS \citep{hu2025automated}
      & \second{55.8} & 40.6 & \second{96.9} & \second{81.2}
      & 60.9 & \second{67.2} & 67.1 \\
    AFlow \citep{zhang2025aflow}
      & 50.0 & \textbf{92.0} & 89.1 & 78.1
      & 57.4 & 62.5 & 71.5 \\
    SwarmAgentic \citep{zhang2025swarmagentic}
      & \second{55.8} & 89.9 & 90.2 & {80.6}
      & \second{62.8} & 53.8 & \second{72.2} \\
    \midrule
    \textbf{\ours}
      & \textbf{58.0} & \second{91.6} & \textbf{99.2} & \textbf{83.6}
      & \textbf{63.3} & \textbf{96.5} & \textbf{82.0} \\
    \bottomrule
  \end{tabular}%
  }
\caption{Held-out test scores (\%) across six evaluation settings
spanning five benchmarks, using Qwen3.5-35B as both optimizer
and executor. Average denotes the unweighted mean of all
\nbench scores. Automated methods receive the same
candidate-evaluation budget (Appendix~\ref{app:baselines}).
Reference architectures are evaluated without search.
Within each section, the \textbf{best} score per column is
bolded and the \second{second-best} is underlined.}
  \label{tab:main_results}
\end{table*}

\definecolor{meanbg}{RGB}{237,233,255}
\begin{table*}[t]
  \centering
  \resizebox{0.88\textwidth}{!}{%
  \setlength{\tabcolsep}{3pt}
  \small
  \begin{tabular}{lrrc>{\columncolor{meanbg}}r}
    \toprule
    \textbf{Method}
      & \textbf{\texttt{gemma4-12B}}
      & \textbf{\texttt{gemma-3-27b}}
      & \shortstack{\textbf{\texttt{Meta-Llama-3.1}}\\\textbf{\texttt{8B-Instruct}}}
      & \textbf{Average} \\
    \midrule
    CoT \citep{wei2022chain}
      & 35.0 & 15.8 & 3.8 & 18.2 \\
    Quality-Diversity \citep{lu2025intelligent}
      & 33.9 & 19.2 & 2.1 & 18.4 \\
    LLM Debate \citep{du2023improving}
      & 43.9 & 20.0 & 4.2 & 22.7 \\
    Reflexion \citep{shinn2023reflexion}
      & 43.9 & 20.8 & 2.9 & 22.5 \\
    CoT-SC \citep{wang2022self}
      & 41.1 & 18.3 & 2.5 & 20.6 \\
    Role Assignment \citep{xu2023expertprompting}
      & 27.8 & 16.7 & 4.2 & 16.2 \\
    \midrule
    ADAS \citep{hu2025automated}
      & \second{51.1} & \second{34.2}
      & \textbf{14.2} & \second{33.2} \\
    AFlow \citep{zhang2025aflow}
      & 42.5 & 19.2 & 3.3 & 21.7 \\
    SwarmAgentic \citep{zhang2025swarmagentic}
      & 49.2 & 20.0 & 2.5 & 23.9 \\
    \midrule
    \textbf{GRAFT}
      & \textbf{66.1} & \textbf{40.0}
      & \second{8.3} & \textbf{38.1} \\
    \bottomrule
  \end{tabular}%
  }
  \caption{ARC test scores (\%) with three executors and the fixed
  Qwen3.5-35B optimizer. Average denotes the
unweighted mean across executors. Evaluation budgets matched. Reference architectures are evaluated
  directly with each executor.}
  \label{tab:executor_comparison}
\end{table*}

\paragraph{Baselines.}
We compare our framework against three automated architecture-search methods: ADAS \citep{hu2025automated}, which generates programs using an archive of prior designs; AFlow \citep{zhang2025aflow}, which performs search with Monte Carlo tree search; and SwarmAgentic \citep{zhang2025swarmagentic}, which uses language-guided swarm optimization. We also report the hand-designed agents that initialize ADAS and
\ours{} (e.g., CoT, CoT-SC, Self-Refine, and LLM Debate, Step-back Abstraction, Role Assignment, Quality-Diversity ), reported and cited in the corresponding results tables, without architecture search explained in Appendix~\ref{app:agents}.

\paragraph{Models.}
We use separate LLMs for architecture optimization and task execution: the optimizer proposes candidate architectures, while the executor performs the model calls specified by those architectures. All models are open-weight and hosted locally, keeping task data, agent outputs, and execution feedback within the evaluation environment throughout the search, an important consideration in privacy-sensitive domains. In the main experiments, we use \texttt{Qwen3.5-35B-A3B-FP8}~\citep{team2026qwen3} as both the optimizer and executor. For the ARC executor study, we fix \texttt{Qwen3.5-35B-A3B-FP8} as the optimizer and compare \texttt{Qwen3.5-9B}~\citep{team2026qwen3}, \texttt{Gemma-4-12B-it}~\citep{team2026gemma}, and \texttt{Gemma-3-27B-it}~\citep{team2025gemma} as executors. On CRUXEval-I, we compare \texttt{Qwen3.5-35B-A3B-FP8} and
\texttt{gpt-oss-120b}~\citep{agarwal2025gpt} as optimizers
with a fixed \texttt{Qwen3.5-9B} executor. Additional executors include \texttt{Qwen3-8B}~\citep{yang2025qwen3} for the case study and \texttt{Llama-3.1 8B}~\citep{grattafiori2024llama} on ARC.
\paragraph{Execution settings.} Our framework maintains five candidate architectures and generates two offspring per generation. Number of generation is specific to each task and is defined as variable. Automated methods are evaluated under matched candidate-evaluation budgets, except SwarmAgentic, which receives a larger budget, using the same task splits, executor, and scoring protocol within each configuration.
Full model, decoding, execution, and budget-accounting settings
are provided in Appendix~\ref{app:baselines}.

\subsection{Results}
\label{sec:main_results}
\paragraph{Overall Performance}
Table~\ref{tab:main_results} reports held-out test accuracy with \texttt{Qwen3.5-35B-A3B} as both optimizer and executor. \ours achieves the highest average accuracy of 82.0\% across the \nbench benchmark settings, outperforming all automated and hand-designed baselines. The largest improvement occurs on SuperGPQA-Hard-Calc, where GRAFT reaches 96.5\% compared with 67.2\% for the strongest baseline on this task. We also observe that \ours remains competitive where hand-designed approaches are strongest. On Game24, LLM Debate outperforms all other automated design baselines, while \ours matches its score of 99.2\%.




\paragraph{Executor choice}
\label{sec:model_sensitivity}
Table~\ref{tab:executor_comparison} reports ARC test performance
with three executor models while keeping the
\texttt{Qwen3.5-35B} optimizer fixed. Hand-designed reference
architectures are evaluated directly with each executor. \ours{} achieves the highest scores with \texttt{gemma4-12B}
and \texttt{gemma-3-27b}, reaching 66.1\% and 40.0\%, respectively.
These results exceed ADAS, the strongest baseline in both
configurations, by 15.0 and 5.8 percentage points. Across the three executors, \ours{} obtains the highest average score of 38.1\%.

\paragraph{Optimizer choice.}
Table~\ref{tab:optimizer_comparison} compares two architecture
optimizers on CRUXEval-I while fixing the executor to
\texttt{Qwen3.5-9B}. \ours{} ranks
first with \texttt{gpt-oss-120b} (60.2\%) and second with \texttt{Qwen3.5-35B} (53.9\%). It is the only automated method to exceed 50\% accuracy with both optimizers, although the relative ranking of search methods changes across configurations.

\begin{table}[t]
  \centering
  \resizebox{\columnwidth}{!}{%
  \begin{tabular}{lrr}
    \toprule
    \textbf{Method} & \textbf{\texttt{Qwen3.5-35B}} & \textbf{\texttt{gpt-oss-120b}} \\
    \midrule
    CoT                   & 33.6 & 33.6 \\
    Quality-Diversity     & 29.7 & 32.0 \\
    LLM Debate            & 40.6 & 34.4 \\
    Reflexion             & 39.8 & 41.4 \\
    CoT-SC                &   32.1 &   36.1 \\
    Role Assignment       &   27.7 &   34.0 \\
    \midrule
    ADAS                  & 51.6 & 24.2 \\
    AFlow                 & 41.4 & \second{50.0} \\
    SwarmAgentic          & \textbf{62.0} & 46.4 \\
    \midrule
    \textbf{\ours}    & \second{53.9} &   \textbf{60.2} \\
    \bottomrule
  \end{tabular}%
  }
  \caption{CRUXEval-I test scores (\%) using two architecture optimizers
  and the fixed Qwen3.5-9B executor. Evaluation budgets are matched.}
  \label{tab:optimizer_comparison}
\end{table}

\begin{figure*}[t]
  \centering
  \includegraphics[width=0.75\textwidth]{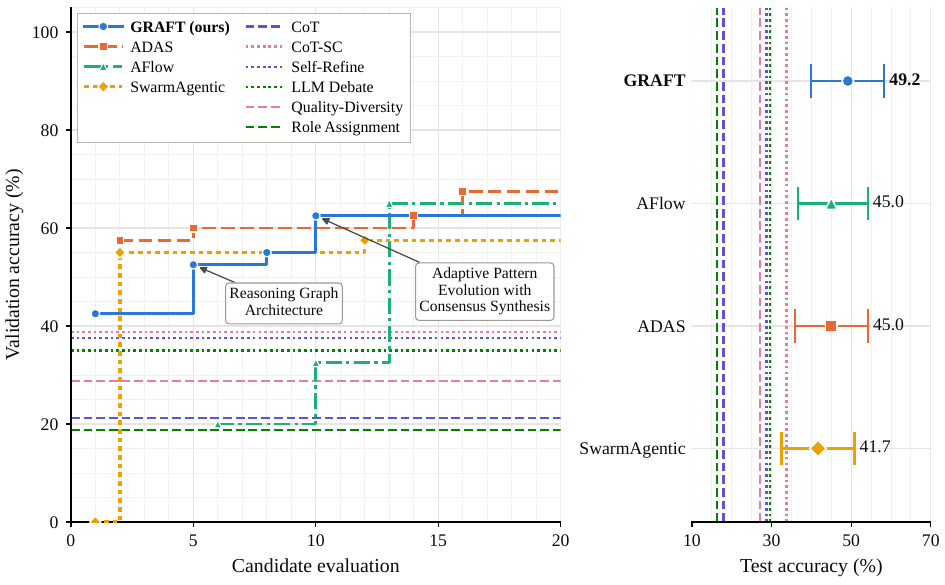}
  \caption{Search on ARC with optimizer \texttt{Qwen3.5-35B} and executor \texttt{Qwen3.5-9B}.
  Left: best validation accuracy found so far versus the
  number of candidate systems evaluated. Right: test accuracy
  of each method's best-validation system, with 95\% bootstrap confidence intervals.
  Dashed lines: hand-designed agents.}
  \label{fig:arc_search}
\end{figure*}

\section{Analysis and Ablation}
\label{sec:analysis}
\label{sec:additional_analyses}

\begin{figure}[!htbp]
  \centering
  \includegraphics[width=0.9\columnwidth]{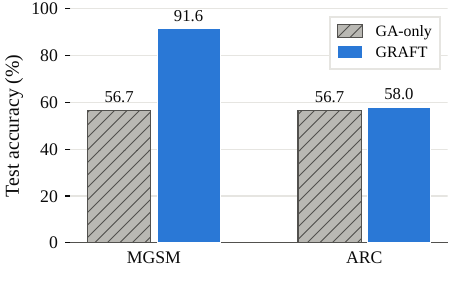}
  \caption{Effect of the behavioral-feedback harness. Test accuracy (\%) of the
    best-by-validation agent found by genetic search alone (GA-only) and by
    \textsc{GRAFT} (genetic search with the harness) on MGSM and ARC, \texttt{Qwen3.5-35B-A3B} as both optimizer and executor.}
  \label{fig:ga-vs-hgad}
\end{figure}

\paragraph{Contribution of Execution Evidence.}

\label{sec:evidence_analysis}

Figure~\ref{fig:ga-vs-hgad} compares \ours with GA-only on MGSM and ARC task. GA-only retains parent code, aggregate fitness, elite context,
and the same crossover and integrated mutation instructions,
but omits per-example execution evidence. Initialization,
selection, replacement, and design budgets are identical
across conditions. We report the test accuracy of the agent
selected by validation performance in each condition. Adding
the harness increases test accuracy from 56.7\% to 91.6\%
on MGSM, a gain of 34.9 percentage points, and from 56.7\%
to 58.0\% on ARC, a gain of 1.3 percentage points.

Our diagnostics suggest that the larger observed improvement
on MGSM reflects how directly execution evidence identifies
actionable program changes. For example repeated error messages may expose
an answer-type mismatch, which high-performing offspring
address by converting answers to formats that are
expected by the evaluator. A single correction can therefore
prevent the same interface failure,
whereas aggregate fitness alone does not identify its shared
cause. This improvement illustrates task-specific interface
adaptation, an important aspect of automated agent design.
Even when the executor does not consistently produce the
expected representation, the surrounding scaffold can enforce
it through explicit normalization.
\paragraph{Search Dynamics and Test Performance.}
Figure~\ref{fig:arc_search} compares ARC search trajectories
and the test performance of the agents selected by validation
accuracy, using \texttt{Qwen3.5-35B} as the optimizer and
\texttt{Qwen3.5-9B} as the executor. The best-so-far validation
accuracy of \ours{} rises from approximately 43\% to 63\%
during the first half of the plotted search and then plateaus. On the test set. \ours{} achieves 49.2\%, compared with 45.0\% for both ADAS and AFlow and
41.7\% for SwarmAgentic. The experiment shows that its validation--test gap is also
the smallest among the compared methods (13.3 percentage points).



\section{Case Study: Execution Evidence on CRUXEval-I}
\label{app:case_study}

\begin{figure*}[t]
  \centering
  \includegraphics[width=.85\textwidth]{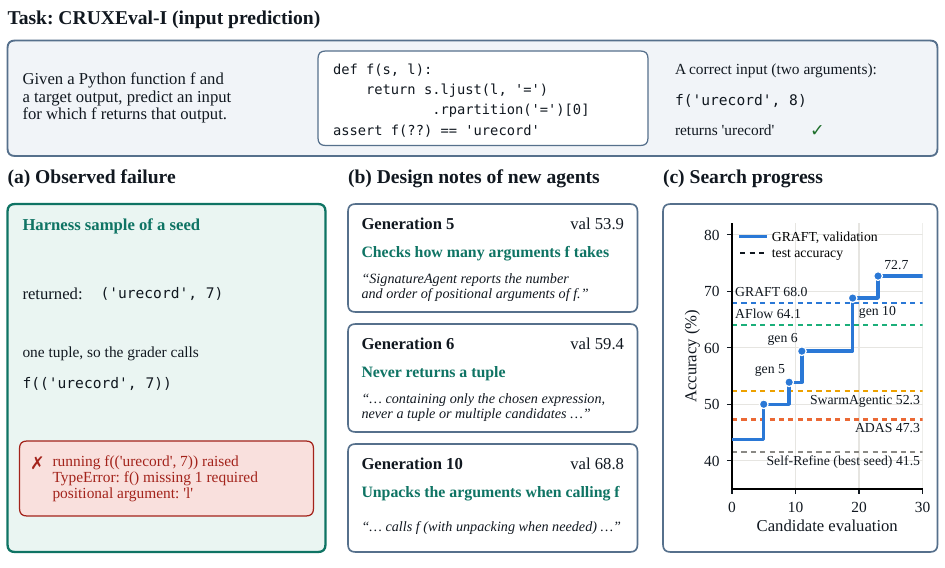}
  \caption{Execution evidence in a GRAFT run on CRUXEval-I
    (optimizer gpt-oss-120b, executor Qwen3-8B).
    (a)~A failure sample from the harness evidence of a hand-designed seed:
    the agent returns both arguments of $f$ as one tuple, so calling $f$
    raises a \texttt{TypeError}.
    (b)~Design notes of offspring from generations 5, 6 and 10, each with a
    short summary, a verbatim excerpt and its validation accuracy.
    (c)~Best validation accuracy so far in each generation. The dashed lines
    mark the test accuracy of best seed and best agents from automated methods.}
  \label{fig:cruxeval_case}
\end{figure*}

Figure~\ref{fig:cruxeval_case} follows one GRAFT run on CRUXEval-I with
\texttt{gpt-oss-120b} as the optimizer and \texttt{Qwen3-8B} as the executor. In this task,
an agent receives a Python function $f$ and a target output and must predict
an input for which $f$ returns that output. For CRUXEval-I, each failed
execution record also contains the evaluator's message, for example the
exception raised when $f$ is called on the predicted input.

\paragraph{Observed failure.}
The hand-designed agents often predicted the right values in the wrong
form: they returned all arguments of $f$ as a single tuple, so calling $f$
raised a \texttt{TypeError} for a missing positional argument
(Figure~\ref{fig:cruxeval_case}a). In every seed that produced execution
records, this error accounted for three to five of the six failed samples
shown to the proposal LLM. Aggregate fitness alone does not reveal this
cause: it only reports that, for example, the best seed solves 43.8\% of the
validation tasks.

\paragraph{Changes in offspring.}
The error remained visible in the harness evidence of the parents selected in
later generations, and the design rationales of their offspring address it
directly (Figure~\ref{fig:cruxeval_case}b). In generation~5, the LLM added an
agent that reports the number and order of $f$'s positional arguments
(validation accuracy 53.9\%); in generation~6, it required the final answer to
contain a single expression and never a tuple (59.4\%); and in generation~10,
it unpacked the predicted arguments before calling $f$ (68.8\%). In the
execution records of the best offspring, the error appears in only one or two
of the six failed samples.

\paragraph{Outcome and limitations.}
The best system, found in generation~12, reaches 72.7\% validation and
68.0\% test accuracy, compared with 41.5\% for the best
hand-designed agent (Figure~\ref{fig:cruxeval_case}c). This example shows
how the proposal LLM uses execution evidence when it is available. In this case study we did not run a
matched GA-only search in this configuration, however we got the test accuracy for SwarmAgent, Aflow and ADAS equal to 52.3, 64.1, and 47.3 respectively with the same candidate-evaluation budget, it is worth mentioning that each generation  with GRAFT equals 2 candidate evaluation since two offspring is generated in each generation.

\section{Conclusion}
\label{sec:conclusion}

\paragraph{Summary.} We introduced \ours{}, which uses aggregate fitness to select parents and survivors, while parent-specific execution feedback guides semantic crossover with integrated mutation. The harness provides bounded samples of successful and failed validation runs, reusing records from fitness evaluation without exposing reference answers or requiring additional task executions. Experiments across reasoning and programming tasks show gains over hand-designed agents and competitive performance against automated architecture-search methods, with rankings varying across optimizer and executor configurations. These findings support combining evolutionary search with execution feedback for automated agent design.



\paragraph{Limitations.}
Our evaluation covers selected reasoning and programming benchmarks
under specific model configurations and search budgets. The results
do not establish performance on open-ended, interactive, or
long-horizon agent tasks. Moreover, evaluating the search framework
across different tasks and models does not demonstrate that a
discovered architecture transfers to new tasks or executors without
further optimization. Held-out test
evaluation addresses performance on unseen instances within the
evaluated setting, but does not establish robustness to distribution
shift. 

\paragraph{Broader Impact and Societal Considerations.} Automated agent design may reduce the engineering effort required
to build task-specific systems and make architecture exploration
more accessible to researchers and developers. Our use of open-weight
models supports reproducibility and local deployment, including
settings where task inputs and execution records need to remain
within controlled infrastructure. Automating architecture discovery could also facilitate the development
of agents for harmful or unauthorized activities. Deployment decisions
should therefore consider misuse risks and performance under realistic
conditions, with human oversight appropriate to the application.



\bibliography{custom}

\appendix

\clearpage
\appendix

\makeatletter
\providecommand{\appendixtablecaption}[2]{%
  \begingroup
  \def\@captype{table}%
  \caption{#1}\label{#2}%
  \endgroup
}
\makeatother

\twocolumn[{
\begin{minipage}{\textwidth}
\section{Benchmark Details}
\label{app:benchmarks}

For each benchmark we fix a validation split, used as the fitness signal
during search, and a disjoint test split, used only to evaluate the final
selected system. All methods use the same splits within a configuration.
Table~\ref{tab:benchmarks} lists the splits and scoring rules.

\medskip
\begingroup
\centering
\small
\setlength{\tabcolsep}{3pt}
\renewcommand{\arraystretch}{1.4}
\begin{tabular}{@{}p{0.21\textwidth}p{0.38\textwidth}p{0.23\textwidth}p{0.06\textwidth}p{0.06\textwidth}@{}}
  \toprule
  \textbf{Benchmark} & \textbf{Task} & \textbf{Score} & \textbf{Val} & \textbf{Test} \\
  \midrule
  ARC & Infer a grid transformation from input--output examples and apply it to a new input & Exact match of the output grid & [128] & [200] \\
  MGSM & Grade-school math word problems in 11 languages & Exact match of the integer answer & [128] & [800] \\
  Game of 24 & Combine four numbers with $+,-,\times,\div$ to obtain 24 & Valid expression equal to 24 & [128] & [400] \\
  MBPP+ & Write a Python function from a natural-language description & Passes the visible and hidden EvalPlus tests & [128] & [247] \\
  SuperGPQA-STEM & Multiple-choice questions in science, engineering, and medicine & Accuracy & [128] & [500] \\
  SuperGPQA-Hard-Calc & Hard calculation questions, disjoint from STEM & Accuracy & [128] & [500] \\
  CRUXEval-I & Predict an input that yields a given output of a Python function & Executed output matches & [128] & [672] \\
  \bottomrule
\end{tabular}
\par
\appendixtablecaption{Benchmark splits and scoring.}{tab:benchmarks}
\endgroup
\end{minipage}
\medskip
}]

\section{Hand-Designed Reference Agents}
\label{app:agents}

The reference agents are the hand-designed programs released with ADAS.
They form the initial population of ADAS and \ours{}, and are evaluated
without search on the test split with the same executor as the searched
systems. Table~\ref{tab:agents} summarizes them.

\medskip
\noindent\begin{minipage}{\columnwidth}
\begingroup
\centering
\small
\setlength{\tabcolsep}{3pt}
\renewcommand{\arraystretch}{1.1}
\begin{tabular}{@{}p{0.30\linewidth}p{0.66\linewidth}@{}}
  \toprule
  \textbf{Agent} & \textbf{Mechanism} \\
  \midrule
  CoT & One call that reasons step by step before answering. \\
  CoT-SC & 5 CoT samples (temperature 0.8); majority vote. \\
  Self-Refine & Answer, critique, and revise. \\
  LLM Debate & 3 agents with different roles debate. \\
  Step-back & First state the relevant principles, then solve. \\
  Quality-Diversity & Generates up to 3 candidate solutions using different approaches, then selects one as the final answer. \\
  Role Assignment & A router picks an expert role; that expert answers. \\
  \bottomrule
\end{tabular}
\par
\appendixtablecaption{Hand-designed reference agents. Roles are task specific (for MGSM, e.g., Math Professor, Grade School Teacher).}{tab:agents}
\endgroup
\end{minipage}


\section{Implementation Details}
\label{app:baselines}

\paragraph{Common protocol.}
All models are open-weight and served with vLLM. Within each
configuration, all methods use the same validation and test splits,
executor, and scoring rule. The candidate-evaluation budget is 30 for most of the experiments, however it could be variable for optimizing and agent for a specific task. Also, for SwarmAgentic's larger budget described below, because of keeping the method's default optimization process.
Each method returns the system with the highest validation
fitness, which is evaluated once optimization finished on the test split.

\paragraph{\ours{}.}
The population holds $N=5$ programs, initialized with the hand-designed
agents. Each iteration selects parents by tournaments of size $k=3$
and proposes $\lambda=2$ offspring, for 15 iterations mostly to use 30 candidate evaluation budget. The harness keeps up to three successful and six failed executions per program,
taken in validation order, with inputs and outputs truncated to a
task-specific length. 

\paragraph{Baselines.}
\textbf{ADAS} uses the released implementation with the same initial
agents, running 30 iterations of archive-conditioned proposal.
\textbf{AFlow} starts from its blank workflow template and runs 30
optimization rounds with task-appropriate operators (e.g., Custom,
ScEnsemble, and Programmer for mathematics). The best-validation
round is tested.
\textbf{SwarmAgentic} uses its released default schedule of five particles and
ten iterations (50 evaluated systems) and returns its global best system. This
exceeds the 30-proposal budget of the other methods.

\vspace{1cm}
\noindent\begin{minipage}{\columnwidth}
\section{Algorithm}
\label{app:algorithm}
Algorithm~\ref{alg:h-gad} summarizes the complete GRAFT search procedure.

\medskip
\small
\gglmxheading{AI Agentic System Search with \mbox{Gradient guided-LMX}, GRAFT}
\label{alg:h-gad}
\noindent\hangindent=1.3em\hangafter=1
\textbf{Require:} Population size $N\geq2$, iteration limit $n$,
proposal slots $\lambda$, tournament size $k$, GG-LMX, and fixed revision,
retry, budget, and stopping policies.\par
\vspace{2pt}
\noindent\begin{tabular}{@{}r@{\hspace{0.5em}}p{\dimexpr\linewidth-3em\relax}@{}}
\gglmxstep{0}{\textbf{Input:} Valid seed population $P_0$ and validation set $D_{\mathrm{val}}$.}
\gglmxstep{0}{\textbf{Output:} Selected system $S_{\mathrm{best}}$.}
\gglmxstep{0}{\textit{// Initialize search}}
\gglmxstep{0}{\textbf{Evaluate:} Evaluate $P_0$; cache each program's fitness $J(S)$ and execution records $R(S)$.}
\gglmxstep{0}{$t\leftarrow0$.}
\gglmxstep{0}{\textit{// Evolutionary search}}
\gglmxstep{0}{\textbf{while} $t<n$, budget remains, and the stopping condition is unmet \textbf{do}}
\gglmxstep{1}{Initialize $Q_t\leftarrow\emptyset$.}
\gglmxstep{1}{Obtain elite context $\mathcal L_t$ from $P_t$.}
\gglmxstep{1}{\textbf{for} $j=1,\ldots,\lambda$, subject to the remaining budget \textbf{do}}
\gglmxstep{2}{\textbf{Select parents:} Choose two distinct parents using fitness-based tournaments of size $k$.}
\gglmxstep{2}{\textbf{Generate offspring:} Obtain $\widetilde S_c$ using GG-LMX (Algorithm~\ref{alg:gg-lmx}).}
\gglmxstep{2}{\textbf{Evaluate:} Evaluate the proposal under the bounded retry policy.}
\gglmxstep{2}{\textbf{if} a valid, evaluated child $S_c$ is obtained \textbf{then}}
\gglmxstep{3}{Cache its final program version, fitness, and execution records.}
\gglmxstep{3}{$Q_t\leftarrow Q_t\cup\{S_c\}$.}
\gglmxstep{2}{\textbf{end if}}
\gglmxstep{1}{\textbf{end for}}
\gglmxstep{1}{\textbf{Elitist replacement:} $P_{t+1}\leftarrow\mathrm{Top}_N(P_t\cup Q_t;J)$.}
\gglmxstep{1}{$t\leftarrow t+1$.}
\gglmxstep{0}{\textbf{end while}}
\gglmxstep{0}{Choose $S_{\mathrm{best}}\in\arg\max_{S\in P_t}J(S)$.}
\gglmxstep{0}{\textbf{Return} $S_{\mathrm{best}}$.}
\end{tabular}
\par\vspace{2pt}\hrule height 0.8pt
\end{minipage}

\end{document}